# For GPT-4 as with Humans:
# Information Structure Predicts Acceptability of Long-Distance Dependencies

Nicole Cuneo[1], Eleanor Graves, Supantho Rakshit[2], Adele E. Goldberg[1]

{nicole.cuneo@, eg5817@, r.supantho@, adele@}princeton.edu

Departments of Psychology[1] and ECE[2]
Princeton University, Princeton NJ 08544 USA

**Abstract**

It remains debated how well any LM is able to understand natural language or even generate reliable metalinguistic judgments. Moreover, relatively little work has demonstrated that LMs can represent and respect subtle relationships between form and function proposed by linguists. We here focus on a particular such relationship established in recent work: English speakers' judgments about the information structure of canonical sentences predicts independently collected acceptability ratings on corresponding "long distance dependency" (LDD) constructions, across a wide array of base constructions and multiple types of LDDs. To determine whether any LM captures this relationship, we probe GPT-4 on the same tasks used with humans and new extensions. Results reveal reliable metalinguistic skill on the information structure and acceptability tasks, replicating a striking interaction between the two, despite the 0-shot, explicit nature of the tasks, and little to no chance of contamination (Studies 1a, 1b). Study 2 manipulates the information structure of base sentences and confirms a causal relationship: increasing the prominence of a constituent in a context sentence increases the subsequent acceptability ratings on an LDD construction. The findings suggest a tight relationship between natural and GPT-4-generated English, and between information structure and syntax, which begs for further exploration.

## Introduction

There have been divergent reports about the extent to which even the largest Language Models (LMs) represent language in ways that parallel humans. Some have argued that LMs struggle with semantic understanding (Danetella et al., 2024; Basmov, Goldberg, & Tsarfaty 2023; Bender and Koller, 2020; Mandelkern and Linzen, 2023). Others have argued that LMs struggle with acceptability judgments (Danetella et al. 2023; although see Hu et al. 2024), particularly if probed explicitly, as metalinguistic skill is required (Hu and Levy 2023).

Rarely have LMs been tested on more subtle aspects of language such as the *information structure* of sentences (Lambrecht, 1994; Halliday 2013). Information structure determines which parts of sentences are "at-issue" (Potts, 2004), which parts are backgrounded, and to what degree. For instance, the propositional content expressed by the sentences (1a) - (1c) is arguably identical, but the information structure differs.

(1a) She drinks whatever is in the glass.
(1b) The glass, she drinks whatever is in it.
(1c) Whatever is in the glass she drinks.

Humans are remarkably sensitive to information structure properties of sentences (e.g., Francis & Michaelis, 2017; Givón, 2020; Ward & Birner, 2019). Moreover, recent work has established that information structure offers insights into certain natural language phenomena widely regarded to be purely syntactic. The current work is motivated by one such case: Several labs have confirmed an important but widely overlooked relationship between human judgments regarding the information structure of sentences and independent acceptability ratings on long-distance dependency (LDDs) constructions. This relationship and the evidence for it is briefly described below.

## Information structure and LDDs in humans

Linguists and psycholinguists have long been captivated by the fact that certain semantic dependencies between non-local elements of an utterance are deemed more acceptable than others (e.g., Hofmeister, & Sag, 2010; Phillips 2006; Ross 1967; Sprouse, 2013). These "long-distance dependency" constructions (LDDs) include *wh*-Qs, topicalization constructions, relative clauses, and cleft sentences.

Constructions that resist combination with LDD constructions are often referred to as "islands" as if constituents cannot "get off" of the island (Ross, 1967). The majority of theoretical linguistic proposals for accounting for islands have suggested purely syntactic constraints (Chomsky, 1973; Grimshaw, 1986).

At the same time, linguists have long been aware that judgments on LDDs often vary even when the syntax is held constant. For instance, Namboodiripad et al. (2022) establishes experimentally that examples like that in (2) are judged less acceptable than that in (3):

(2) Who did the custodian unlock the door while admitting?
(3) Who did the custodian unlock the door by admitting?

One type of explanation for gradient judgments of LDDs is that unacceptability arises from the combination of constructions with incompatible functions. More specifically, LDD constructions make the constituent in a non-canonical position more prominent, while certain other constructions background the same content to varying degrees (Ambridge & Goldberg, 2008; Cuneo & Goldberg, 2023; Dabrowska, 2013; Erteschik-Shir 1973; Goldberg, 2006, 2013; Lu, Pan, and Degen, 2024; Namboodiripad et al., 2022). In *wh*-Qs like (2) - (3) the *wh*-word (*what*) is prominent, while the causally related adjunct phrase "by admitting" is less backgrounded than the temporal adverb, "while admitting." The claim is essentially that anomaly results when a speaker chooses to foreground and background the same content.

> **Backgrounded Constructions are Islands (BCI):** Constructions are islands to long-distance dependency constructions to the extent that their content is backgrounded within the domain of the long-distance dependency construction (Goldberg, 2006, 2013).

For instance, Namboodiripad et al. (2022) investigated verb-phrase adjuncts, by collecting data on two tasks (see also Ambridge & Goldberg, 2008). One was a Negation task: Following a negated declarative sentence, participants were asked the extent to which the key construction was negated. For example, consider the example stimuli in (4) and (5). Separate groups of participants judged how likely the answer to the question was "yes" after being told to assume the negated sentences in either (4) or (5) was true.

(4) She didn't learn about the Civil War while reading the news. → Did she read the news? [probably yes]
(5) She didn't learn about the Civil War by reading the news. → Did she read the news? [probably not]

The negation task relies on the observation that when an entire sentence is negated, information that is presupposed or taken for granted (i.e. backgrounded) is not affected (Karttunen, 1974), although Backgroundedness is a gradient rather than binary distinction.

Evidence for this idea comes from recent experimental studies using English-speaking human participants (Ambridge & Goldberg, 2008; Cuneo & Goldberg, 2023; Namboodiripad et al., 2022; Lu, Pan, and Degen, 2024). Closely related information structure proposals that emphasize the role of focus rather than prominence provide convergent support that information structure of base sentences plays a key role in acceptability ratings on LDDs (Liu et al. 2022; Mao et al., 2024; Winckel et al., 2025).

In addition to the information structure task, new groups of participants rated the acceptability of corresponding *wh*-Qs (e.g., examples [2] and [3]) or the declarative sentences (without the negation). As predicted by the BCI, constituents that were deemed more prominent (less Backgrounded) were judged more acceptable in corresponding *wh*-Qs, while Backgroundedness played little to no role on the acceptability of the declarative sentences.

Authors (2023) conducted a large-scale experiment with the same design, involving a wide variety of base sentence types (which included a variety of adjuncts, manner-of-speaking verb complements, double objects, to-dative constructions, main clauses, relative clauses, parasitic and non-parasitic gaps). Three types of long-distance dependency constructions were created by hand on each of 144 base sentences (*wh*-Qs, discourse-linked information questions, and relative clauses). Results confirmed that the degree of Backgroundedness robustly predicted acceptability ratings on each type of LDD: more backgroundedness constituents were less available for LDDs.

Lu, Pan, and Degen (2024) provide empirical evidence that the BCI plays a causal role in acceptability judgments of manner-of-speaking verb complements (e.g., *grumble, whisper*), by manipulating the information structure of a context sentence before requesting acceptability ratings on subsequent *wh*-Qs. Specifically, they varied whether the to-be-queried constituent in a *wh*-Q was made prominent in the preceding context, using lexical emphasis. Results demonstrated that acceptability increased after participants had read a context sentence that emphasized the to-be-queried constituent. Fergus et al. (2025) establish the same effect in *wh*-Qs from single conjuncts. Thus, human acceptability ratings on LDDs are causally influenced by the manipulation of information structure in the prior context.

## Motivating the current work

Here was ask, can any LM accurately and explicitly assess the information structure of sentences? If so, do those assessments predict the LM's metalinguistic acceptability ratings on a range of LDDs, mirroring human judgments? Will the same LM also be sensitive to the manipulation of information structure? Since our interest lies in whether this very high bar is met by any LM, we focus here on testing the largest model available: GPT-4 (as of Jan, 2025).

GPT-4 produces and responds to English so naturally that yesterday's science fiction has become today's reality. For this reason, some might argue that GPT-4 *must* appreciate information structure of sentences and which LDDs are acceptable, or its productions and comprehension of English would be noticeably impaired. Yet others have been skeptical about whether LMs do more than parrot phrases from its training (Bender et al., 2021). By probing for evidence of a complex interaction between form and function, the current work may reveal limits on even the most sophisticated of current language models: GPT-4. In fact, despite its vast training data, intricate layers, and remarkable parameter count, there are compelling reasons to question whether GPT-4 will succeed on interaction of tasks tested here.

Since GPT-4 is not open-source, we rely on explicit metalinguistic assessments of information structure and acceptability judgments. It is not clear whether GPT-4 possesses the requisite metalinguistic skills to provide reliable judgments, especially when requiring responses on ordered Likert scales. As Hu and Levy (2023:2) note, "prompting implicitly tests a new type of emergent ability — metalinguistic judgment — which has not yet been systematically explored as a way to evaluate model capabilities." They recommend instead using open-source models which allow the comparison of log conditional probabilities of minimal pairs (see, Gauthier et al. 2020), in order to isolate acceptability from other factors recognized to influence judgments including word frequency, plausibility, and complexity.

Additionally, GPT-4's understanding of information structure may be superficial or unreliable. Indeed, Sieker and Zarrieß (2023) find that BERT models perform poorly on certain types of implicatures related to a speaker choosing one word or construction rather than another (see also Jeretic et al. 2020). Sravanthi et al. (2024) provide a recent benchmark for pragmatic tasks including one that gauges how well various models capture presuppositions. Even the largest model they test (GPT 3.5) performed poorly with only 45% accuracy.

Like humans, LMs find local dependencies easier to process than long-distance dependencies (LDDs) (Kallini, et al., 2024; Warstadt et al., 2020). At the same time, LMs have been found to mirror human judgments of acceptability of LDDs in certain cases including: LDDs from complex nouns in Dutch (Mulders & Ruys, 2024); the PiPPs construction in English (e.g., *wise though he is*) (Potts, 2023); relaxed LDD constraints in Norwegian (Kobzeva et al. 2023). But once again, the question of whether LMs' more general successes rely on superficial characteristics of LDDs is debated (Howitt, et al., 2024).

Note that our quarry extends beyond asking whether human acceptability ratings are mirrored by GPT-4: We first test whether GPT-4's explicit judgments about the information structure of base sentences correlate with their independent acceptability ratings on 4 types of LDDs, more than on ratings of the base sentences themselves. Study 1a finds strong evidence for the predicted interaction and strong correlations with human judgments on the same stimuli. Study 1b replicates the finding using new stimuli to mitigate any contamination effects. Study 2 manipulates the information structure properties of (non-LDD) base sentences and demonstrates that GPT-4's explicit acceptability ratings are causally influenced.

## Study 1a: GPT-4 replicates human data

In order to systematically investigate GPT-4's sensitivity to information structure and acceptability ratings, we deployed a zero-shot, structured prompting framework designed to ensure controlled evaluation, minimize stochastic variation, and prevent contamination from any kind of extraneous model state dependencies. By conducting probes in a zero-shot setting, we ensured the model was not given any prior in-context examples or demonstrations before making judgments. We formulated the prompts in a syntactically constrained, semantically minimalistic format, ensuring that responses were purely model-internal generalizations rather than retrieval-based outputs.

To mitigate response variance due to sampling stochasticity, we queried GPT-4 10 times per unique stimulus to determine reliability as well as mean ratings. The model temperature was set to zero to enforce maximum determinism and minimize soft probability-based variation in scoring. We executed all prompts in a programmatic inference pipeline leveraging the OpenAI API to ensure experimental reproducibility. Each prompt was processed individually with randomized ordering to prevent sequential bias. API requests were throttled to avoid potential caching artifacts from previous completions.

### Acceptability Ratings

GPT-4 was prompted to provide acceptability judgments for the same stimuli previously gathered from human participants (Cuneo & Goldberg, 2023). These included 144 items that allowed for LDDs from verbs with complement clauses, a range of adjuncts, parasitic and non-parasitic gaps, double object and prepositional dative constructions, and transitive clauses and relative clauses. Judgments on base sentences (non-LDDs) and each of 4 types of LDDs were collected: *wh*-Qs, it-clefts, discourse-linked questions, and relative clauses. GPT-4 acceptability ratings were prompted by the text provided in Table 1. All GPT-4 completions were obtained using OpenAI's GPT-4 API between January 27th and January 29th, 2025, with temp. = 0. Model version info was current as of Jan 2025.

The model was instructed to output only a numerical response to prevent contamination from unwarranted verbosity. See Table 1 for an example prompt.

Table 1: Acceptability rating prompt for GPT-4 (Jan 2025)

| |
|---|
| Is the following an acceptable question in English? |
| Who did Marcus write a letter? |
| Rate it on a scale of 1-7, where 1 is very unnatural and 7 is very natural. Return only an integer rating. |

### Backgroundedness Judgments

The same negation task used in prior work to measure the degree of Backgroundedness was also used to probe GPT-4. A sample prompt is provided in Table 2.

Table 2: Example prompt for Backgroundedness judgment provided to GPT-4 (Jan 2025)

| |
|---|
| Assume the sentence below is true and think about what it means: |
| Marcus didn't write her a letter. |
| Now answer the following question with an integer between 1 and 5, where 1 means no, 2 means probably not, 3 means can't tell, 4 means probably yet, and 5 means yes. |
| Did Marcus write her a letter? |

**Results** An ordinal model was created with acceptability ratings as outcome, sentence type (base or LDD) and Backgroundedness as fixed interacting effects, and the maximal random effect structure convergence allowed. As expected, GPT-4 rated LDD sentences less acceptable than base sentences overall ($\beta$ = -8.96, $z$ = -3.70, $p$ < .0001). The predicted interaction is significant ($\beta$ = -1.82, $z$ = -2.84, $p$ = .004): Increased backgroundedness predicts lower scores each of the LDD constructions. A comparison with human judgments and acceptability ratings is provided in Figure 1.

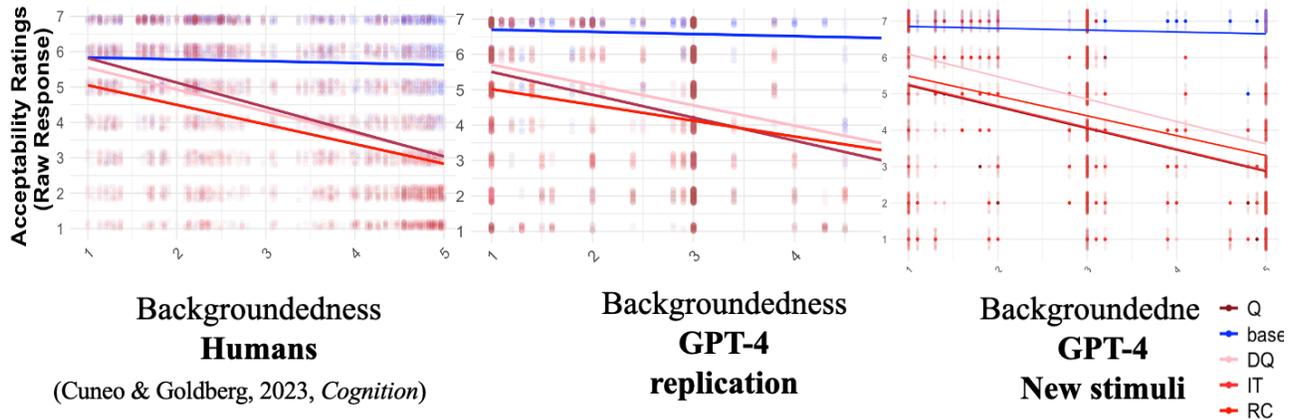

Figure 1: Comparison between human data [Left panel]; GPT-4 on the same stimuli [middle]; GPT-4 on newly created stimuli [Right]. Raw backgroundedness judgments on base sentences (x-axis) and Acceptability ratings on each LDD (shades of red) and base sentences (blue) (y-axis). Data available here: https://researchbox.org/4231

**Results** Using the same ordinal model as in Study 1a, results again show that Backgroundedness judgments on base stimuli predict acceptability ratings on LDDs more than they predicted ratings on the base sentences themselves. That is predicted interaction is significant ($ß = -.29$, $z = -9.80$, $p < .00001$), as shown in Figure 1 (rightmost panel).

## Study 2: Manipulating Backgroundedness

A second study manipulates Backgroundedness (or conversely, the prominence) of a constituent on the same data collected in Study 1b. The method was adapted from Lu, Pan, and Degen (2024) who confirmed a causal role of information structure on *wh-Q*s in the case of manner-of-speaking verbs (See also Fergus et al. [submitted] for the manipulation on single conjunct extraction). A context sentence is provided before an acceptability rating is requested on a following *wh-Q*. In half of trials, the context sentence emphasizes the to-be-queried content with lexical stress (Emphasis condition); in the No-emphasis condition, the same context sentences are provided without lexical stress, followed by acceptability ratings on the same set of *wh-Q*s. Here we test GPT-4 on this manipulation by explicitly prompting it to "Please focus on the part in ALL CAPS," in the Emphasis condition. Example prompts are provided in Table 4.

Table 4: Example trial structure used in Experiment.

| Condition: | **Emphasis** | **Non-emphasis** |
|---|---|---|
| Context sentence: | Sam: Marcus didn't write HER a letter. | Sam: Marcus didn't write her a letter. |
| *Wh-*Q | Chris: Then who did Marcus write a letter? | |
| Acceptability Rating Prompt: | Is Chris' question acceptable in English? Rate it on a scale of 1-7 where 1 is very unnatural and 7 is very natural. Return only an integer rating. | |

**Results** To systematically test for an influence of emphasis in the preceding context sentence, a linear mixed-effects model was applied mean acceptability ratings. Since 50 % of the *wh-Q*s received mean acceptability ratings of 6 or higher on a scale of 7-point scale, we examined the items that received ratings in the non-emphasis condition of less than 6 ($n = 72$) The model predicted mean ratings, with Emphasis (emphasis vs. no emphasis) as a fixed factor. Random intercepts for item and construction type were included Results reveal that context-sentences that included emphasis resulted in significantly higher ratings on the *wh-Q*s that followed compared to context sentences without emphasis. ($ß = 0.44$, $t = 8.48$, $p < .00001$). Results are provided in Figure 2.

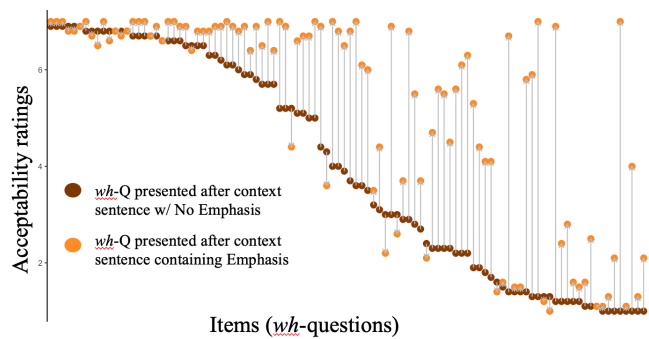

Figure 2: GPT-4's mean acceptability ratings of *wh-*Qs after a context sentence without emphasis (dark) and b) with emphasis on the to-be-queried constituent (light orange).

## General Discussion

The current work establishes that GPT-4 captures systematic relationships between information structure and syntactic acceptability, despite the 0-shot, explicit nature of the tasks, and little to no chance of contamination. Results replicate a striking interaction between these two factors only recently confirmed in English-speakers.

Specifically, Study 1a replicates a study on humans (Cuneo and Goldberg, 2023) and establishes that GPT-4 reliably assesses information structure on canonical (base) sentences; moreover, those judgments predict its independent acceptability ratings on three types of long-distance dependency constructions: *wh-Q*s, discourse-linked questions, and relative clauses. That is, results confirm that for GPT-4, as for humans, more backgrounded constituents are less available for LDDs.

Study 1b replicates Study 1a, using novel stimuli to address contamination concerns, including a new LDD construction (*it*-clefts) not tested in the original study on humans. Highly converging results were again found. Together these findings indicate that GPT-4's ability to process and evaluate LDD constructions is not an artifact of memorization, but instead reflects a relationship between information structure and the acceptability of corresponding long-distance dependencies, despite the latter widely presumed to be syntactic (since Chomsky, 1973).

Critically, a second study strengthens our claim that GPT-4's metalinguistic acceptability ratings on LDDs are influenced by the information structure of the preceding context. Study 2 manipulated the information structure of base sentences just before being prompted for acceptability ratings on *wh-Q*s. Results demonstrate a causal influence of the information structure in the prior context on acceptability ratings. These findings align with recent human studies on complements of manner-of-speaking verbs (Lu, Pan, & Degen, 2024) and on single-conjunct extraction (Fergus et al., submitted). That GPT-4 is sensitive to this type of manipulation indicates that it captures subtle and *flexible* generalizations in natural language.

Time and space constraints prevent us from attempting to compare and contrast the current findings with work that has argued that even GPT-4 fails to understand presuppositions (Basmov, et al., 2023) or inferences (Dentella et al., 2024) and fails to provide accurate acceptability judgments (Dentella et al., 2023). We note, however, that the bar we set was determined by a full-scale psycholinguistic experiment, which collected non-binary judgments from hundreds of naïve human participants. The current work also collected judgments and ratings on ordered Likert scales, providing only the simplest possible instructions to GPT-4. Effects of context are expected to influence people, and Study 2 demonstrates that context influences GPT-4 as well.

## Limitations and Future Directions

Our ultimate goal as cognitive scientists is to understand how *humans* learn and represent subtle relationships between form and function that emerge in natural languages. The current work indicates that at least the largest LM, GPT-4, captures such a relationship and flexibly adapts in context (recall Study 2). Since GPT-4 is a proprietary model, future work should replicate these findings using open-weight models (e.g., Mixtral, LLaMA-3) and log-prob-based scoring techniques. to determine whether they show the same flexible and robust relationship between information structure and syntax reported here. This would also offer the opportunity to vary or ablate aspects of the training data to explore how LMs capture the requisite generalizations.

We also recognize that GPT-4's judgments in no way substitute for direct empirical studies of human cognition. Future work needs to probe the limits of the BCI by testing new base constructions, new types of LDDs and non-English (and non-WEIRD) languages.

## Conclusion

We are sympathetic with the observation that if GPT-4 didn't respond to language in a way that was extremely similar to people, it would be patently obvious, as was the case just a few short years ago. Yet quantitative results are required to compare behavior between humans and LMs. Here we establish for the first time that GPT-4 robustly captures a subtle and flexible *relationship* between the information structure of base sentences and acceptability ratings on different sentences that contain long-distance dependencies.

Studies 1a, 1b, and 2 demonstrate that GPT-4 is capable of representing and applying a systematic relationship between information structure and syntactic acceptability, paralleling human judgments across diverse constructions. These findings suggest that, despite its lack of explicit training, or hard-coded rules, GPT-4 exhibits an emergent understanding of information structure's role in syntactic acceptability. The inclusion of new stimuli in Study 1b, including a new LDD construction (*it*-clefts) and the emphasis manipulation in Study 2 demonstrate the model's ability to generalize beyond prior training data, to flexibly but robustly capture a non-obvious relationship between gradient judgments of presuppositions on one set of sentences to acceptability judgments on other types of sentences, whose acceptability is widely presumed to be syntactic.

In bridging the divide between linguistic theory and artificial intelligence, our findings reveal an unexpected yet striking parallel between human cognition and GPT-4's metalinguistic competence. The model not only aligns with human judgments on the interplay between information structure and long-distance dependencies but also adapts flexibly to context—a hallmark of linguistic generalization. That GPT-4 successfully internalizes and applies a complex, non-trivial interaction between information structure and acceptability suggests that its capacity extends beyond mere memorization or shallow heuristics. While the precise mechanisms remain opaque, these results underscore the potential of large language models as tools for linguistic inquiry, both as subjects of study and as instruments for probing the underlying structure of human language. As AI continues to evolve, so too will the questions we ask of it—questions that, as this work demonstrates, can yield insights not only about machines, but about the very nature of human language itself.

2025. *Proceedings of the Cognitive Science Society.*
## Acknowledgements

We are grateful to Arielle Belluck for editing a penultimate version of the manuscript, and to Princeton's Natural and Artificial Minds program for funding.